\documentclass{IOS-Book-Article}

\usepackage{mathptmx}
\usepackage{times}
\usepackage{epsfig}
\usepackage{graphicx}
\usepackage{amsmath}
\usepackage{amssymb}
\usepackage{multirow}
\usepackage[table,xcdraw]{xcolor}
\usepackage{caption}
\usepackage[draft]{todonotes}
\usepackage{xcolor}
\usepackage{url}
\usepackage{subfig}
\newcommand\norm[1]{\left\lVert#1\right\rVert}
\def\hb{\hbox to 10.7 cm{}}

\begin{document}

\pagestyle{headings}
\def\thepage{}

\begin{frontmatter}              % The preamble begins here.

%\pretitle{Pretitle}
\title{Retinal Optic Disc Segmentation using Conditional Generative Adversarial Network}

\markboth{}{May 2018\hb}
%\subtitle{Subtitle}

\author[A]{\fnms{Vivek Kumar } \snm{Singh}%
\thanks{Corresponding Author:  E-mail:
vivekkumar.singh@urv.cat}},
\author[A]{\fnms{Hatem A.} \snm{Rashwan}},
\author[D]{\fnms{Farhan } \snm{Akram}},
\author[B,C]{\fnms{Nidhi} \snm{Pandey}},
\author[A]{\fnms{Md. Mostafa Kamal} \snm{Sarker}},
\author[A]{\fnms{Adel} \snm{Saleh}},
\author[A]{\fnms{Saddam} \snm{Abdulwahab}},
\author[A]{\fnms{Najlaa} \snm{Maaroof}},
\author[A]{\fnms{Jordina Torrents} \snm{Barrena}},
\author[A]{\fnms{Santiago} \snm{Romani}}
and
\author[A]{\fnms{Domenec} \snm{Puig}}
\runningauthor{B.P. Manager et al.}
\address[A]{Department of Computer Engineering and Mathematics,  Universitat Rovira i Virgili, 43003 Tarragona, Spain.}
\address[B]{Kayakalp Hospital, 110084 New Delhi, India.}
\address[C]{Sant Joan de Reus University Hospital, 43204 Reus, Spain.}
\address [D]{Imaging Informatics Division, Bioinformatics Institute, 30 Biopolis Street, $\#$ 07-01 Matrix, 138671, Singapore.}

\begin{abstract}
This paper proposed a retinal image segmentation method based on conditional Generative Adversarial Network (cGAN) to segment optic disc. The proposed model consists of two successive networks: generator and discriminator. The generator learns to map information from the observing input (i.e., retinal fundus color image), to the output (i.e., binary mask). Then, the discriminator learns as a loss function to train this mapping by comparing the ground-truth and the predicted output with observing the input image as a condition. 
Experiments were performed on two publicly available dataset; DRISHTI GS1 and RIM-ONE. The proposed model outperformed state-of-the-art-methods by achieving around $0.96$ and $0.98$ of Jaccard and Dice coefficients, respectively. Moreover, an image segmentation is performed in less than a second on recent GPU.
\end{abstract}

\begin{keyword}
Conditional Generative Adversarial Networks, deep learning, retinal image analysis, optic disc segmentation
\end{keyword}
\end{frontmatter}
\markboth{May 2018\hb}{May 2018\hb}

\section{Introduction}

Retinal fundus image analysis is very important for doctors to deal with the medical diagnosis, screening and treatment of opthalmologic diseases. The morphology of the optic  disk (OD), which is a location, where ganglion cell axons exit the eye to form the optic nerve in which visual information of the photo-receptors is transmitted to the brain, is an important structural indicator for assessing the presence and severity of retinal diseases, such as diabetic retinopathy, hypertension, glaucoma, hemorrhages, vein occlusion, and neovascularization \cite{macgillivray2014retinal}. Retinal OD segmentation is the first step for a significant investigation of retinal images which helps to cause eye diseases \cite{almazroa2015optic}. 
%\todo{references added as \cite{paperlabel} you should make a bibfile in your project}

\begin{figure}[htp]
\centering
\includegraphics[width=0.35\textwidth, height=0.2\textheight]{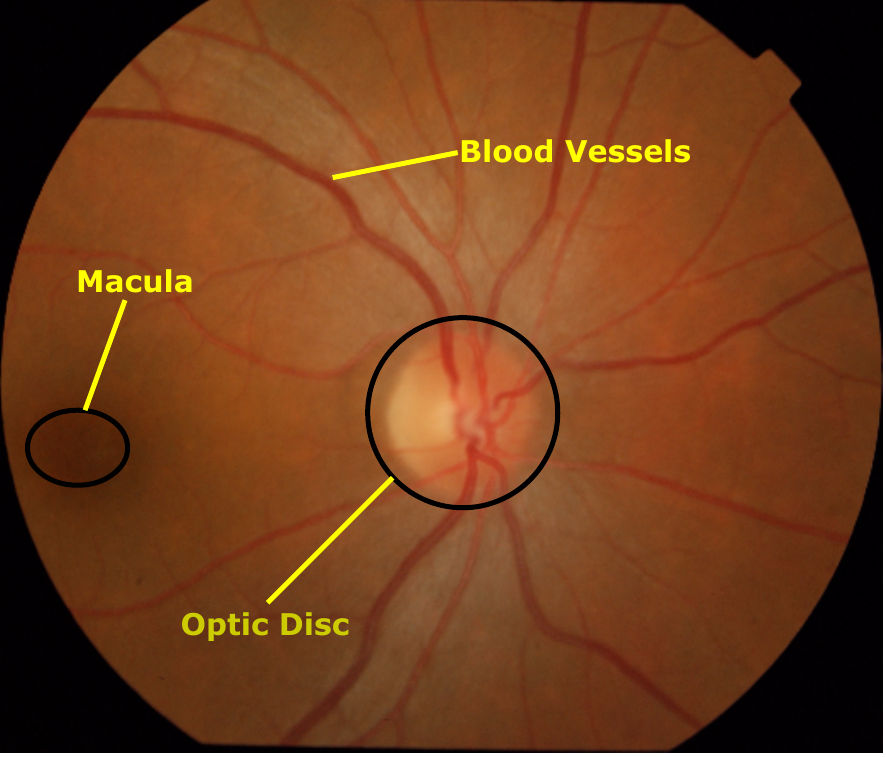}
\caption{Relevant structures in a fundus image.}
\label{fig:figD}
\end{figure}

The OD appears as a bright yellowish oval region within color fundus images through which the blood vessels enter the eye. The macula is the center of the retina, which is responsible for our central vision. Figure~\ref{fig:figD} shows the color retinal fundus image with the key anatomical structures denoted. For ophthalmologists and eye care specialists, an automated segmentation and analysis of fundus optic disc plays an important role to diagnose and treat the retinal diseases. 

Numerous methods has been proposed to detect and segment the optic disc. For diagnosis of glaucoma disease, Chrastek et al.\cite{chrastek2005automated} proposed an automated segmentation algorithm to segment the optic nerve head. They firstly removed the blood vessel by using a distance map algorithm and a morphological operation, and then anchored active contour model has been used to segment the optic disc. Lowell et al. \cite{lowell2004optic} proposed a deformable contour model to segment the optic nerve head boundary of retinal images by using a template matching and a directionally sensitive gradient to discard the interference of vessels. In turn, Welfer et al.\cite{welfer2010segmentation} proposed an automated optic disk segmentation in a fundus image using an adaptive morphological operation. They then used a watershed transform marker to define the optic disk boundary. In addition, the vessel obstruction is minimized by morphological erosion. 

With the increase of using deep learning models in segmentation tasks, many methods have recently been proposed based on  convolutional neural network (CNN). An automatic optic disc and cup image segmentation has been proposed in \cite{al2018multiscale} based on a stack of deep U-Net models. Each model in a cascade refines the result of the previous one. 

In this paper, we propose a retinal OD segmentation model based on conditional Generative Adversarial Network (cGAN) \cite{isola2017image}. cGANs is a deep learning network that can learn the statistical invariant features (texture, color etc.) of input image and segment the optic disc region. This paper introduces, to the best of our knowledge, the first application of the  conditional generative adversarial training for retinal optical disc segmentation. The Proposed cGAN network consists of two combined networks: generator and discriminator. The generator network learns the mapping from the input, a fundus image, to the output, a segmented image. In turn, the discriminator (i.e, adversarial term) learns a loss function to train this mapping by comparing the ground-truth and the predicted output. Finally, the whole cGAN network optimizes a loss function that combines a conventional binary cross-entropy loss with an adversarial term. The adversarial term encourages the generator to produce output that cannot be distinguished from ground-truth ones.

The rest of the paper is organized as follows. Firstly, section 2 describes the methodology of the proposed cGAN model. In addition, section 3 shows the experiments and discussion. Finally, the conclusion and some future lines of research are explained in section 4.

\section{Proposed Methodology}

\begin{figure}[htp]
\centering
\includegraphics[width=0.9\textwidth, height=0.7\textheight]{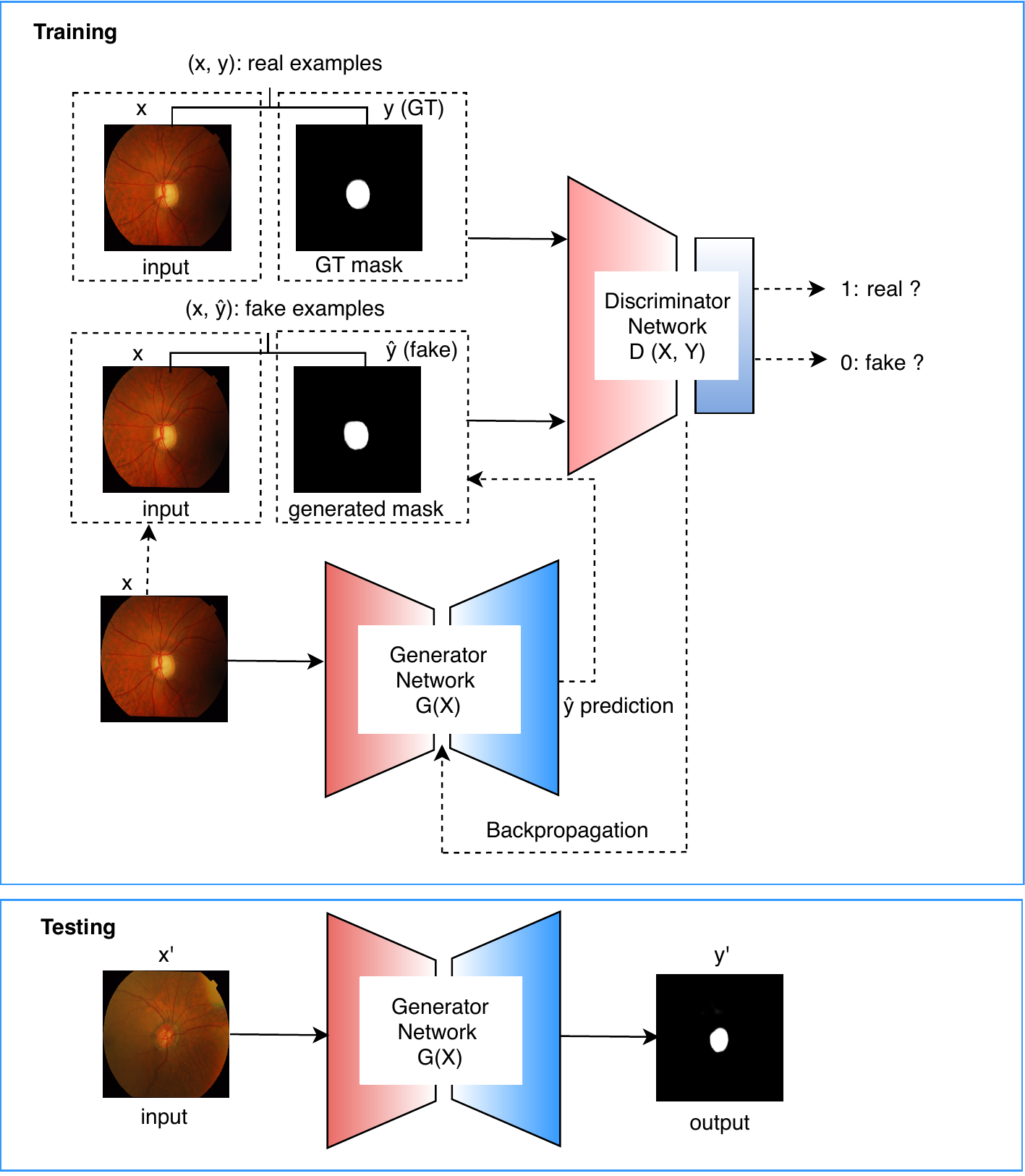}
\caption{General framework for optic disc segmentation.}
\label{fig:figmodel}
\end{figure}

Figure~\ref{fig:figmodel} shows the proposed cGAN framework for optic disc segmentation model. Optic disc detection in this paper is addressed as a segmentation problem, which is carried out by the generator network. 

\begin{figure}[htp]
\centering
\includegraphics[width=\textwidth, height=0.4\textheight]{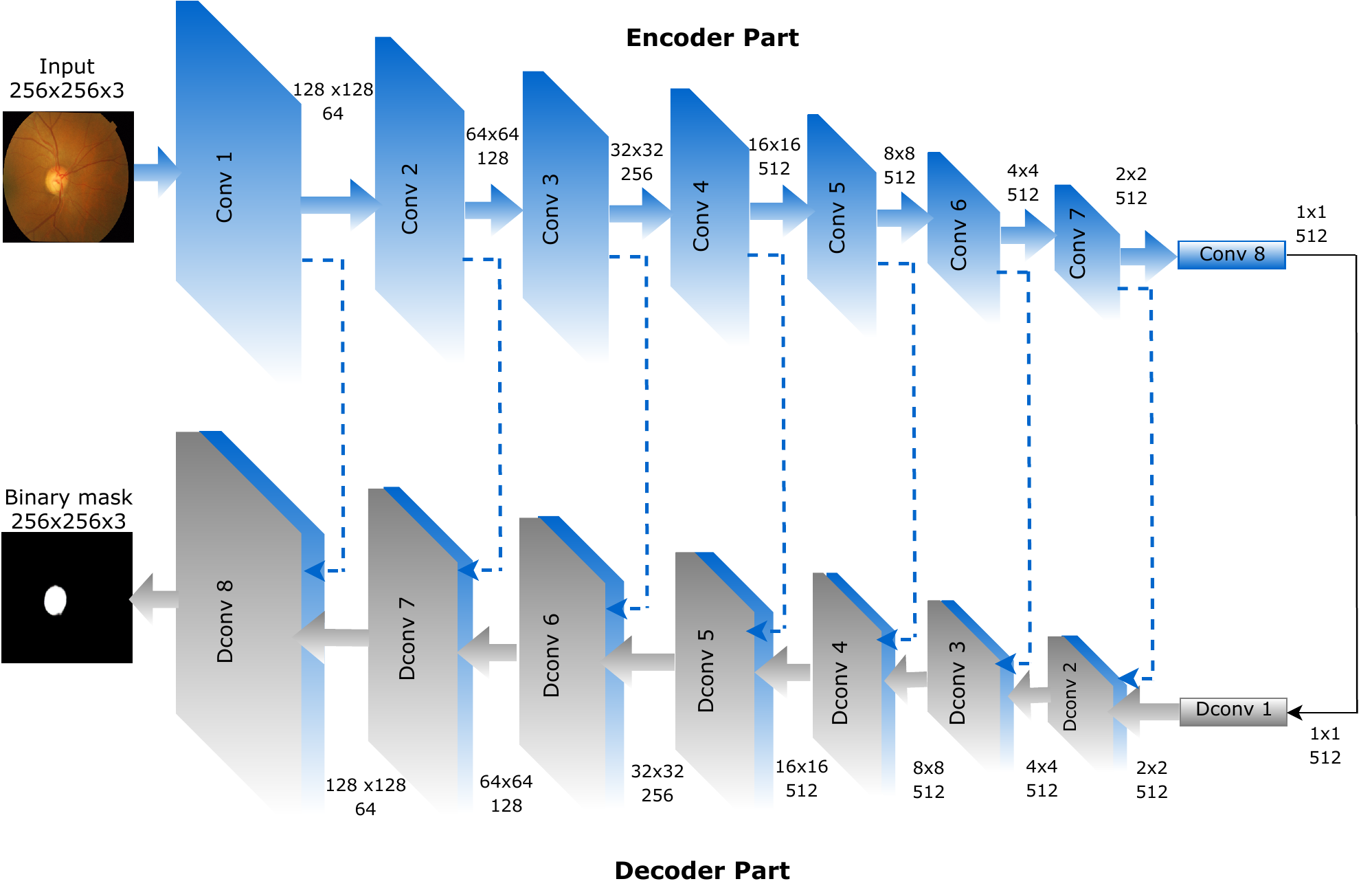}
\caption{ The generator network architecture, composed by layers of encoder part and decoders part}
\label{fig:figGenerator}
\end{figure}

The generator network is based on encoding and decoding layers. The function of encoders network is to extract features from the input retinal fundus images by covolutional filters with down-sampling, in turn, the decoders utilized the decovolutional filters with up-sampling the feature maps to predict the final segmented image. Each (de)covolutional layer is followed by batch normalization.We used LeakyRelu activation function with slope 0.2 in the end of each (de)covolutional layer. The size of the each spatial filter in each convolution and deconvolution is 4x4 to down- and up-sample the feature maps size with a stride $2\times 2$. At the last convolutional layer in encoders, Tanh activation function is used. In the last layer of the decoders, we used a sparse fully connected layer (FC), which convert the feature maps into a single dimensional vector and using sigmoid activation function for obtaining the binary class optic disc segmentation. 

In order to increase the segmentation performance of the proposed network, skip connections are used (shown in dotted lines in Figure 3) between encoders and decoders by concatenating the feature maps of a convolutional layer with the ones resulted from the corresponding deconvolutional layer. The main advantage of skip connection is that encoder learn the high level features of retinal optic disc image pixels and decoder learn to correlate with the encoder features to determine whether following receptive fields of the output image are likely to belong to the optic disc mask or not. In Figure~\ref{fig:figGenerator}, the generator network architecture is shown, which consists of encoder and decoder layers. As an input, a retinal image is observed. The original input images from both two publicly dataset are very big in size, therefore in order to reduce the network size we resized the input image to a $256\times 256$ size and the value of each pixel is normalized to [0,1].

The architecture of the discriminator, which observes the concatenation of the retinal image and the segmentation mask as an input to be evaluated as real or fake, is composed of five convolutional layers. Thus, including the adversarial score in the loss computation of the generator fosters its capabilities to provide good segmentation. Each convolution layer used $3\times 3$ spatial filter with a stride $2\times 2$. The first layer of the discriminator generates $64$ feature maps extracted from the input image. In turn, the second and third layers produce $128$ and $256$ feature maps respectively. The fourth layer generates $512$ feature maps with a $30\times 30$ output size. 

Suppose $x$ and $y$ showing a retinal fundus image and corresponding ground truth segmentation with random variable $z$ respectively. $G(x, z)$ is the predicted binary mask of the optic disc. Besides, the L1 normalized distance between ground truth and the predicted masks is $y-G(x, z)$. In addition,  $\lambda$ is an empirical weighting factor and the discriminator output score is $D(x, G(x, z))$. If the discriminator output score gives 1 then predicted mask seems like a true ground truth, otherwise it gives output score 0. 

\begin{equation}
\ell_{Generator} (G,D) = E_{x,y,z} \big(-log (D(x,G(x,z)))\big)+ \lambda E_{x,y,z} \big(\norm{y-G(x,z)}_{1}\big), 
\end{equation}

Here, we have used a L1 loss function to boost the learning process. According to [7], using only L1 loss will produce blurred segmentations. Therefore, to avoid this problem, we used adversarial network to increase the performance of the segmented image. Adversarial network allows the generator to completely change the output image at fine level. 
The loss computation of discriminator network is shown below:

\begin{equation}
\ell_{Discriminator} (G,D) = E_{x,y} \big(-log (D(x,y))\big)+ E_{x,y,z} \big(-log (1-D(x,G(x,z)))\big)
\end{equation}

Therefore, the optimizer helps to the discriminator network in order to maximize the belief value for actual masks (by minimizing $-log (D(x, y))$ and to minimize the belief value for generated masks (by minimizing $-log (1-D(x, G(x, {y})))$.

We have used the Adam \cite{kingma2014adam} optimizer with learning rate 0.0002 for optimization. In addition, during experiment, the batch size is set to 4 and the model is trained with 200 epochs. The proposed cGAN model permits an accurate and strong learning even with a few number of hundreds training images . After segmentation, we have applied erode and dilation morphological operations to remove some white artifacts from the generated output binary mask.

\section{Experiments and Discussion}
In the experiments, we used a 64-bit I7-6700, 3.40GHz CPU with 16GB of memory space as well as NVIDIA GTX 1070 GPU, running on Ubuntu 16.04 Linux operating system. We used Pytorch \cite{paskze2017tensors} neural network library to devise a neural network model using the deep learning framework.

We conduct a comprehensive set of experiments to validate the
potential of our proposed model on two datasets:

%\subsection{Datasets}
\textbf{DRISHTI-GS1 \cite{sivaswamy2015comprehensive}}: The dataset is publicly available with comprises 101 images, which is divided into a training and a testing set of images. Training and testing sets consist of 50 and 51 images respectively. These images have their corresponding binary mask as ground truth.

\textbf{RIM-ONE \cite{fumero2011rim}}: This dataset is publicly available particularly for Optic nerve head segmentation, it has a total of 169 high resolution images with their corresponding ground truth. We have used 100 images as training and rest 69 images for test purpose.

For quantitative assessment of the performance of OD segmentation, we have computed Accuracy, Dice Coefficient (Dice), Jaccard index (JACC), Sensitivity and Specificity as detailed in Table~\ref{Table1}. We have performed the experiments using the two datasets with three common segmentation methods, FCN \cite{long2015fully}, U-Net\cite{ronneberger2015u} and SegNet \cite{badrinarayanan2017segnet}. In addition, we compared our results with three baseline state-of-the-art methods, such as Shankaranarayana  et. al. \cite{shankaranarayana2017joint}, Maninis et. al. \cite{maninis2016deep} and Zilly et. al. \cite{zilly2015boosting}. %\cite{}. %\todo{please here each method with its reference. I removed GAN-Unet}

%\subsection{Experimental results}
% \usepackage{multirow}
\begin{table}[htb]
\centering
\caption{Accuracy, Dice coefficient, Jaccard index, Sensitivity and Specificity with the cGAN ,FCN, SegNet and Unet models, in addition to three baseline methods evaluated on DRISHTI GS1 and  RIM-ONE. The best results are marked  in a bold text. Results for optic disc segmentation (-) indicates that the result is not reported.} 
\label{Table1} 
\scalebox{0.9}{
\begin{tabular}{|c|c|c|c|c|c|c|}
\hline
Methods                      & Dataset         & Accuracy        & Dice            & JACC            & Senstivity      & Specificity     \\ \hline
\multirow{2}{*}{FCN}         & DRISHTI GS1            & $0.93$          & $0.91$          & 0.89          & 0.92          & 0.96          \\ \cline{2-7} 
                             & RIM-ONE & 0.94          & 0.92          & 0.87          & 0.88          & 0.95 \\ \hline
                             
\multirow{2}{*}{SegNet} & DRISHTI GS1             & 0.94          & 0.88 & 0.83 & 0.89 & 0.95          \\ \cline{2-7} 
                             & RIM-ONE & 0.93          & 0.85          & 0.78          & 0.86          & 0.94 \\ \hline

\multirow{2}{*}{U-Net} & DRISHTI GS1             & 0.97          & 0.95 & 0.90 & 0.96 & 0.98          \\ \cline{2-7} 
                             & RIM-ONE  & 0.94          & 0.92          & 0.89          & 0.93          & 0.97          \\ \hline

 \multirow{2}{*}{Shankaranarayana  et al.2017 \cite{shankaranarayana2017joint}}    & DRISHTI GS1             & \textbf{-} & - & - & \textbf{-} & \textbf{-} \\ \cline{2-7} 

                             & RIM-ONE  & -          & 0.98          & 0.88          & -          & -          \\ \hline

\multirow{2}{*}{Maninis et al. 2016 \cite{maninis2016deep}}    & DRISHTI GS1             & \textbf{-} & \textbf{-} & \textbf{-} & \textbf{-} & \textbf{-} \\ \cline{2-7} 

                             & RIM-ONE  &          & 0.96          & 0.89          & -          & -          \\ \hline

\multirow{2}{*}{Zilly et al. $2016$ \cite{zilly2015boosting}}    & DRISHTI GS1             & \textbf{-} & 0.97 & 0.91 & \textbf{-} & \textbf{-} \\ \cline{2-7} 

                             & RIM-ONE  & \textbf{-} & 0.94          & 0.89          & -          & -          \\ \hline

\multirow{2}{*}{\textbf{cGAN} (our proposed)}      & DRISHTI GS1             & \textbf{0.98}          & \textbf{0.97 }         & \textbf{0.96 }         & \textbf{0.98  }        & \textbf{0.99 }         \\ \cline{2-7} 

                             & RIM-ONE  & \textbf{0.98}          & \textbf{0.98}          & \textbf{0.93}          & \textbf{0.98}          &\textbf{ 0.99 }         \\ \hline
\end{tabular}}
\end{table}

In this section, the proposed method is evaluated on: DRISHTI GS1 and RIM-ONE datasets
to show its robustness in a comparison to the state-of-the-art methods. 
%We have trained our model on train and test on DRISHTI GS1 dataset and evaluated on RIM-ONE dataset.

Table~\ref{Table1} shows that quantitative results of the performance of our proposed segmentation method on using both publicly available DRISHTI GS1 and RIM-ONE datasets. As shown, with DRISHTI GS1, the cGAN model can segment the OD regions with around $98\%$, $97\%$, $96\%$, $97\%$ and $99\%$ of Accuracy, Dice coefficient, Jaccard index, sensitivity and specificity, respectively. As well as, the proposed method outperformed the six tested segmentation models. However, the U-net model provided acceptable results and comparable to our results with with around $97\%$, $95\%$, $90\%$, $96\%$ and $98\%$ with the five evaluation matrices. The three tested baseline methods have only computed the Dice coefficient and Jaccard index as shown in Table~\ref{Table1}. The work proposed in \cite{zilly2015boosting} yielded feasible scores with around $97\%$ and $91\%$ of the dice coefficient and Jaccard index, respectively.

Furthermore, in order to support out the aforementioned results, we evaluated our model on RIM-ONE dataset. The resulted Accuracy, Dice coefficient, Jaccard, sensitivity and specificity scores with our model were achieved around $98\%$, $98\%$, $93\%$, $97\%$ and $99\%$, respectively. With RIM-ONE, our proposed cGAN model also outperformed the compared approaches in terms of the six evaluation metrics. \cite{shankaranarayana2017joint} yielded the best Dice coefficient among the five tested methods with around $98\%$, in turn the model in \cite{zilly2015boosting} provided the best among the five tested methods. The method proposed by Shankaranarayana\cite{shankaranarayana2017joint}, the dice coefficient $98\%$ is equal to our proposed method, however our cGAN model achieved high Jaccard index of $93\%$ as comparison to $88\%$. In addition, The U-net model has still provided good results comparing the other semantic segmentation methods.

\begin{figure*}
\centering
\includegraphics[width=0.9\textwidth, height=0.4\textheight]{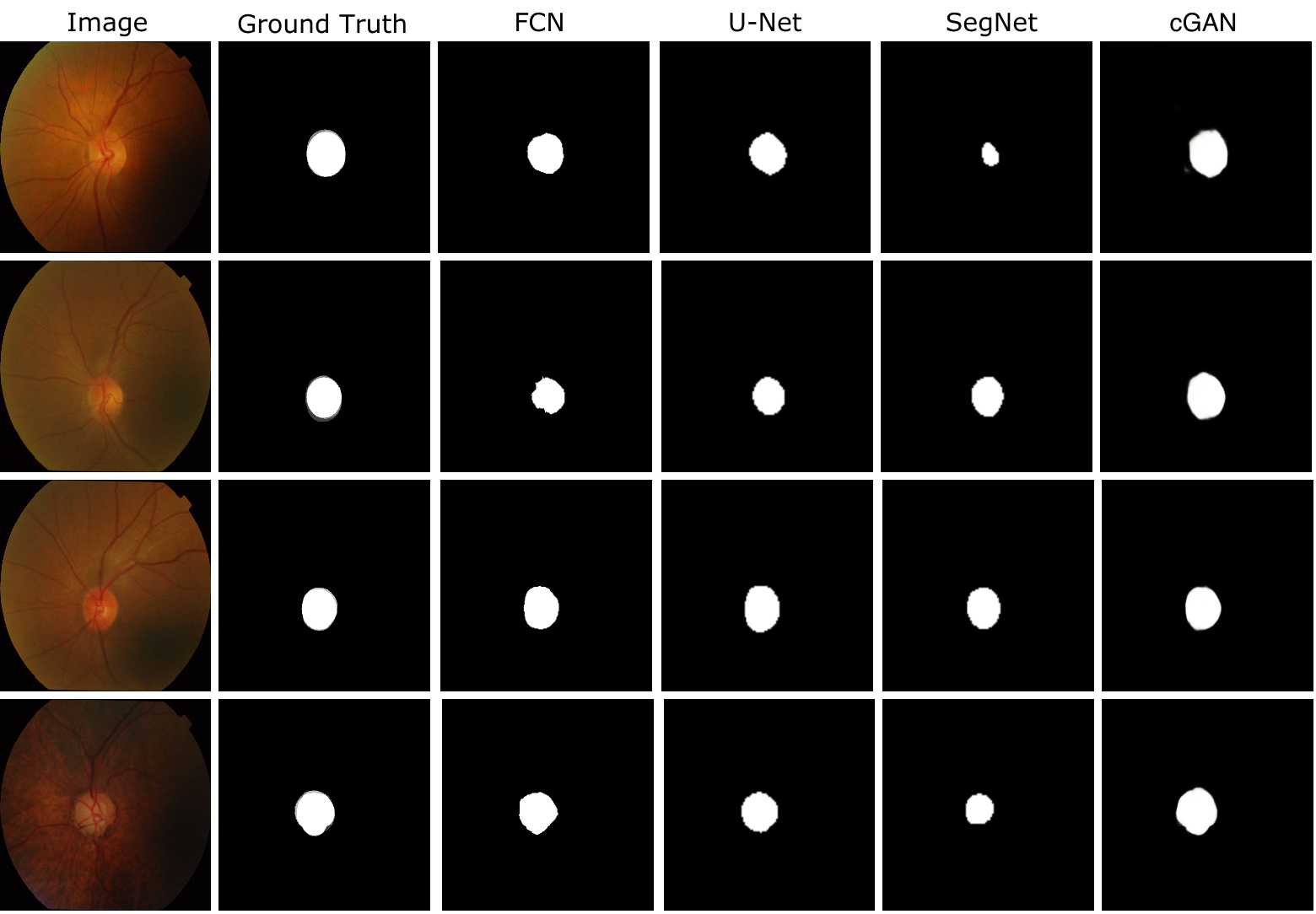}
\caption{ Examples of the retinal optic disc segmentation : (col 1) retinal optic disc images, (col 2) ground-truth masks, (col 3) FCN, (col 4) U-Net, (col 5) SegNet and (col 6) generated masks with the cGAN. 
}
\label{figresults}
\end{figure*}

A qualitative comparison of segmentation results with the state-of-the-art methods using both retinal optic disc datasets is shown in Figure~\ref{figresults}. As shown, the OD segmentation with the proposed method is much closer to the ground truth with accurate boundaries compared to results of the state-of-the-art methods. The visualization supports our numerical results and The U-Net also provided acceptable segmentation. In turn, the SegNet yielded the worst segmentation among the five tested methods.

%\todo{add results of one common segmentation model and on the baseline methods with explanation}

\section{Conclusion}
This work proposed a deep learning framework based on conditional Generative Adversarial Network (cGAN) to segment the retinal fundus optic disc. The cGAN consists of two networks: generator and discriminator. To train properly, the cGAN network does not require a large number of images to train. In addition, it renders a high segmentation performance without adding any complexity, since the final segmentation is only achieved with the generator network. Experimental results show that the cGAN outperformed the state-of-the-art optic disc segmentation methods. Future work will aim at validating our approach on more and larger datasets, which confirm to the opthalmology into a clinical practice. In addition, using  our proposed model in a comprehensive diagnosis model for practically analyzing fundus images.

\bibliographystyle{ios1}
\bibliography{my_bib}

\end{document}